\documentclass[10pt,journal,compsoc]{IEEEtran}
\usepackage{xcolor}
\usepackage[nottoc,notlof,notlot]{tocbibind} % Put the bibliography in the ToC
\ifCLASSOPTIONcompsoc
  \usepackage[nocompress]{cite}
\else
  \usepackage{cite}
\fi
\ifCLASSINFOpdf
  \usepackage[pdftex]{graphicx}
\else
  \usepackage[dvips]{graphicx}
\fi
\usepackage{amsmath}
\usepackage[ruled,vlined]{algorithm2e}
\usepackage{float}
\usepackage{stfloats}
\begin{document}
\title{ExamGAN and Twin-ExamGAN \\ for Exam Script Generation}

\author{Zhengyang Wu, 
        Ke Deng, %~\IEEEmembership{Fellow,~OSA,}
        Judy Qiu,
        Yong Tang% <-this % stops a space
\IEEEcompsocitemizethanks{\IEEEcompsocthanksitem Zhengyang Wu, Yong Tang(corresponding author 2)  are with the School of Computer Science, South China Normal University, Guangzhou, 510631, China.\protect\\
E-mail: {wuzhengyang}@m.scnu.edu.cn,
{ytang}@m.scnu.edu.cn

\IEEEcompsocthanksitem Ke Deng (corresponding author 1) is with the School of Science, RMIT University, Melbourne, VIC3001, Australia.\protect\\
 Email: {ke.deng}@rmit.edu.au% <-this % stops an unwanted space

\IEEEcompsocthanksitem Judy Qiu is with the Business School, University of Western Australia, Perth, WA6009, Australia.\protect\\
 Email: {judy.qiu}@uwa.edu.au}% <-this % stops an unwanted space

\thanks{%Manuscript records.
© 2021 IEEE. Personal use of this material is permitted. Permission from IEEE must be obtained for all other uses, in any current or future media, including reprinting/republishing this material for advertising or promotional purposes, creating new collective works, for resale or redistribution to servers or lists, or reuse of any copyrighted component of this work in other works. }}

\IEEEtitleabstractindextext{%
\begin{abstract}
Nowadays, the learning management system (LMS) has been widely used in different educational stages from primary to tertiary education for student administration, documentation, tracking, reporting, and delivery of educational courses, training programs, or learning and development programs. Towards effective learning outcome assessment, the exam script generation problem has attracted many attentions and been investigated recently. But the research in this field is still in its early stage. There are opportunities to further improve the quality of generated exam scripts in various aspects. In particular, two essential issues have been ignored largely by existing solutions. First, given a course, it is unknown yet how to generate an exam script which can result in a desirable distribution of student scores in a class (or across different classes). Second, while it is frequently encountered in practice, it is unknown so far how to generate a pair of high quality exam scripts which are equivalent in assessment (i.e., the student scores are comparable by taking either of them) but have significantly different sets of questions. To fill the gap, this paper proposes ExamGAN (Exam Script Generative Adversarial Network) to generate high quality exam scripts, and then extends ExamGAN to T-ExamGAN (Twin-ExamGAN) to generate a pair of high quality exam scripts. Based on extensive experiments on three benchmark datasets, it has verified the superiority of proposed solutions in various aspects against the state-of-the-art. Moreover, we have conducted a case study which demonstrated the effectiveness of proposed solution in a real teaching scenario. 
\end{abstract}

\begin{IEEEkeywords}
Educational Data Mining, Exam Script Generation, Generative Adversarial Network, Deep Knowledge Tracing
\end{IEEEkeywords}}

\maketitle

\IEEEdisplaynontitleabstractindextext

% \fi
\IEEEpeerreviewmaketitle

\IEEEraisesectionheading{\section{Introduction}\label{sec:intro}}

\IEEEPARstart{T}he learning management system (LMS) has been widely applied in different educational stages from primary to tertiary education. Exams is an important part of learning outcome assessment. Manually generating high quality exam scripts by teachers is a confronting task physically and intelligently which requires a thorough understanding of the knowledge points covered by each question, the relative importance of knowledge points in the course, the academic performance distribution of students in the class, and the proper difficulty level, etc. This task becomes particularly challenging in MOOCs (Massive Open Online Courses) where the number of courses, classes, and students are typically much larger than those in traditional schools and the size of question banks are considerably large. 

This motivates the automatic generation of exam script in the past decade. In \cite{kamya2014fuzzy}, questions are randomly selected from different question databases without bias where each database contains questions of a particular type. In \cite{chavan2016apriori}, the questions have been extracted from a question bank against a predetermined difficulty level (e.g., the average of student scores is 70 if 100 is full). In \cite{Yuan2012examination}, the ability of exam script to distinguish academic performances between students has been studied. In \cite{yildirim2010Genetictpg}, author has developed a model based on a genetic algorithm so as to optimize objectives including proper difficulty level and clear distinction between students in academic performance. In \cite{El-Rahman2019Shuffling}, randomization algorithm has been adopted in exam script generation with various goals including proper distribution of question types and knowledge points. 

The research in this topic is still in its early stage. There are opportunities to further improve the quality of generated exam scripts in various aspects. In particular, two issues have been largely neglected by existing solutions. First, given a course, it is unknown yet how to generate an exam script which can result in a desirable distribution of student scores in a class (or across different classes \footnote{The rest of the paper only mention a class of students for simplicity.}), i.e., a normal distribution where the mean can implicate the difficulty level and the standard deviation can disclose the capability to distinguish academic performances between students. Second, it is often encountered in practice two exam scripts are required, for example, one for formal exam and the other for deferred exam. However, it is unknown so far how to generate a pair of high quality exam scripts which are equivalent in assessment but have significantly different sets of questions. Two exam scripts are equivalent in assessment if student scores are comparable no matter which script is taken. 

To fill the gap in this research field, we propose to generate high quality exam scripts featured by \textit{validity} (i.e., the proper knowledge coverage), \textit{difficulty} (i.e., the expected average score of students), \textit{distinguishability} (i.e., the ability of distinguishing academic performances between students), and \textit{rationality} (i.e., the desirable student score distribution). These goals motivates us to develop \textit{ExamGAN} (Exam Script Generative Adversarial Network). Given a class of students in a course, the generated exam scripts are optimized to (i) properly cover the course knowledge points and (ii) have proper student score distribution with desirable mean (i.e., difficulty level) and standard deviation (i.e., ability to distinguish academic performances between students). In specific, the student scores on an exam script are estimated using \textit{Deep Knowledge Tracing} \cite{piech2015DKT} by exploring the exercise records of students. Different from \cite{piech2015DKT}, instead of predicting the probability that a student correctly answers a question, we are interested in the score distribution of all students in the class. ExamGAN is based on conditional \textit{Generative Adversarial Network} (GAN) model \cite{mirza2014cgan} where the conditions are the unique features of students in a class.    

Moreover, ExamGAN has been extended in this paper to \textit{Twin-ExamGAN} (or simply \textit{T-ExamGAN}) which consists of two ExamGAN models, each generating one exam script. While exam scripts are optimized by the corresponding ExamGAN individually to achieve high quality, T-ExamGAN provides the mechanism to enforce both exam scripts with significantly different sets of questions but they are equivalent in assessment. Two learning strategies have been proposed for different optimization priorities. 

In summary, the contributions of this study are threefold:
\begin{itemize}
    \item This study has proposed the ExamGAN model to automatically generates high quality exam scripts which have demonstrated unique advantage in various aspects.     
    \item This study has proposed the T-ExamGAN model which, for the first time to the best of our knowledge, generates a pair of high quality and equivalent exam scripts, and enforces the two exam scripts with significantly different sets of questions. 
    \item While applying GAN (Generative Adversarial Network) in exam script generation for the first time, a series of deliberately designed techniques have been proposed and tailored to meet the inimitable requirements of ExamGAN and T-ExamGAN. The effectiveness has been verified via substantial tests on benchmark datasets.   
\end{itemize}

Table \ref{tbl:annotation} introduces the notations used in this study. The rest of the paper is organized as follows. We briefly review the related work in Section \ref{sec:related}. In Section \ref{sec:pre}, we introduce the preliminaries. After that, Section \ref{sec:model} proposes the ExamGAN and Section \ref{sec:model2} presents the T-ExamGAN. The experimental evaluations are reported in Section \ref{sec:test} and a case study is presented in Section \ref{sec:casestudy}. Finally, this study is concluded in Section \ref{sec:conclusion}.

 \begin{table}
  \caption{A summary of notations.}
  \label{tbl:annotation}
  \centering
  \begin{tabular}{c||p{6.5cm}}
    \hline
    Notation & Description\\
    \hline
    $\Lambda$ & A course \\ \hline
    $QuB$ & The exam question bank of course $\Lambda$ \\ \hline
    $ExB$ & The set of exercises in course $\Lambda$ \\ \hline
    $q\in QuB$ & A question in $QuB$ \\ \hline
    $e\in ExB$ & A question in $ExB$ \\ \hline
    $K$ & The knowledge points of course $\Lambda$ \\  \hline
    $K_q$ & The knowledge points covered by question $q\in QuB$ \\ \hline
    $K_e$ & The knowledge points covered by exercise $e\in ExB$ \\ \hline
	$w(k)$ & The weight of knowledge point $k\in K$ in course $\Lambda$ \\ \hline
	$\lambda\in \Lambda$ & A class of course $\Lambda$  \\ \hline
	$E$ & An exam script for class $\lambda$  \\ \hline
	$S$ & The students in class $\lambda$ \\ \hline
	$R(s,E)$ & The score of student $s\in S$ on exam script $E$ \\ \hline
	$<e, a_s>$ & $e\in ExB$ is an exercise answered by student $s\in S$ and $a_s$ is whether the answer is correct or not \\ 
	\hline
\end{tabular}
\end{table}

\section{RELATED WORK}\label{sec:related}
%\subsubsection*{Educational Data Mining}
%With the extensive application of online learning systems, online course platforms, and online learning management systems, various types of educational data have been collected on a daily basis. Educational Data Mining (EDM) is a research area that uses data mining techniques in an educational environment to serve purposes such as improving the learning outcomes, the student performance estimation, and the learning process optimization, which benefit learners, educators, and administrators \cite{romero2007educational,baker2009state}. 

%In \cite{bakhshinategh2018educational,romero2010educational}, authors divide the application of EDM into three parts, namely \textit{student modeling}, \textit{decision support systems}, and \textit{other applications}. The tasks of student modeling include predicting student performance, detecting undesirable student behaviors and so on. The decision support systems aim at providing visualization reports, generating recommendation, constructing course-ware, and more. Course-ware is defined as educational software that provides content, video, testing, and other learning materials. The purpose of building course-ware is to help teachers automatically create or develop course materials using student information. Obviously, the core of building course-ware is to select and organize learning materials from existing resources.

\subsubsection*{Learning Materials Selection}
%Effectively selecting and organizing learning materials in a large learning platform and making the use of these resources predictable is the driving force behind the continuous development of online learning activities. 

An emotion-based learning resource recommendation architecture has been proposed where, beyond personal information, the learners' preferences, behaviors \cite{rajendran2018behavior}, and emotions \cite{tiam2019emotions} are explored. In \cite{christudas2018evolutionary}, it measures the learner's satisfaction during the learning process according to the learner's knowledge mastery level and the interaction behavior, and adjusts the organizational learning materials accordingly. In \cite{10.5555/3298239.3298437}, the question difficulty is predicted.
In \cite{zhu2018mt} \cite{10.1145/3379507} \cite{8744302}, authors have proposed to model the knowledge mastery level based on the questions answered so far which can be used in personalized test question recommendation. In \cite{mongkhonvanit2019deep}, authors have organized the interactive information generated by learners in a timely manner to predict the learner's response to the next task. Knowledge tracing \cite{corbett1994knowledge} predicts the probability that a learner can answer a given question correctly. In \cite{piech2015DKT}, authors have proposed an improved knowledge tracing model based on the exercises answered, called \textit{Deep Knowledge Tracing} (DKT), using LSTM network to predict the probability to correctly answer the next exercise.   

\subsubsection*{Automatic Exam Script Generation}
Given a question bank, the automatic exam script generation aims to select a subset of questions from the question bank to form an exam script. In \cite{bobad2018estudy}, it first specifies a predetermined difficulty level and then extract different questions from a question bank based on an algorithm. According to \cite{LiuXP2003statistics}, the desirable difficulty level should be near 0.7, that is, the average score of all students is 70 if 100 is full. In \cite{kamya2014fuzzy}, it tries to improve the objectivity of the difficulty level of the questions in exam scripts such that the test results are close to the specified difficulty level. In \cite{chavan2016apriori}, authors have introduced a model using Apriori algorithm to match the difficulty level of the selected questions according to the difficulty level set by the examiners.

In addition to difficulty level, the distinguishability (aka. discrimination) has been taken into account in exam script generation. It refers to the ability to distinguish academic performances between students. According to \cite{Yuan2012examination}, the distinguishability is measured by $\frac{P_H-P_L}{100}$ where $P_H$ is the average of the $27\%$ highest scores and $P_L$ is the average score of the $27\%$ lowest scores. In \cite{yildirim2010Genetictpg}, author has developed a model based on a genetic algorithm such that the problem of generating exam scripts is transformed into a multi-objective optimization problem. The optimization objectives include difficulty and distinguishability.  

Moreover, the distributions of question types and/or knowledge points in the exam script have been considered. In \cite{kamya2014fuzzy}, authors have proposed a system which utilizes a fuzzy logic algorithm. The purpose is to randomly select questions from different question databases without bias where each database contains questions of a particular type such as objective questions or subjective questions. In \cite{El-Rahman2019Shuffling}, authors have used a shuffling algorithm as a randomization technique with the attempt to avoid repetitively selecting same questions for the same examinees. In \cite{chim2019randomization}, authors have used the randomization algorithm to select the attributes of questions (such as difficulty, type, etc.) or parameters (such as corresponding knowledge points, scores, etc.) to generate exam scripts. In \cite{saikia2019aptitude}, authors have proposed to automatically generates aptitude-based questions with certain keywords using randomization technique. 

In short, automatic exam script generation is still in its early stage and there are opportunities to further improve the quality of generated exam scripts in various aspects. Particularly, two essential issues are largely neglected by existing studies as discussed in Section \ref{sec:intro}.

\subsubsection*{Generative Model}
The proposed solution ExamGAN is based on conditional Generative Adversarial Network (GAN) which is a generative model. With generative models, it assumes data is created by a probability distribution, which is then estimated and a distribution very similar to the original one is generated. The generative models include Gaussian Mixture Models (GMM), Hidden Markov Models (HMM), Latent Dirichlet Allocation (LDA), Boltzmann Machines (RBM), Variational autoencoders (VAE), and Generative Adversarial Network (GAN) \cite{GM2020100285}.  

When the GMM fits on a data, it is a generative probabilistic model which can generate new data distributed similar to the distribution to which the GMM was fit \cite{4218101}. A GMM assumes all the data points are generated from a mixture of a finite number of Gaussian distributions with unknown parameters. The HMM are used to generate sequences named as Markov chains. LDA is a generative technique mainly used for topic modelling, although more broadly it is considered a dimensionality reduction technique \cite{944937}. Restricted Boltzmann machines (RBM) represent a class of unsupervised neural networks which generate data to form a system that closely resembles the original system \cite{105555}. All these four models cannot support the exam script generation investigated in this study. 

Normal autoencoders comprise three layers. The input layer $X$, the middle layer $Z$ (aka., coding layer) and the output layer $\hat{X}$. The inputs are encoded into feature extracted representations and stored in $Z$ through weights. Similarly, an output similar to $X$ at $\hat{X}$ is generated after decoding of vector $Z$. In order to generate data, VAEs assume that $X$ is generated by a true prior latent distribution $Z$, which is assumed to be a Gaussian. In addition to $Z$, the conditional VAEs consider more information relevant to the generation \cite{doersch2016tutorofVAE}. While VAEs generate data similar to the original data to an extent, GANs \cite{an2015variational} can achieve the higher accuracy in generating data \cite{GM2020100285}. GANs offer a solution of training generative models without the procedure of maximizing a log likelihood (as VAEs) which are usually intractable and require numerous approximations. This motivates us to generate exam scripts based on GANs in this study while the conditional VAEs can also be used to generate exam scripts.

\section{Preliminary}\label{sec:pre}
Given a course denoted as $\Lambda$, the knowledge points covered by the course is denoted as $K$. For each knowledge point $k\in K$, $w(k)$ denotes the importance of $k$ in the course. The exam question bank in the course is denoted as $QuB$ and the set of exercises is denoted as $ExB$. The knowledge points covered by a question $q\in QuB$ is $K_q$. Similarly, the knowledge points covered by an exercise $e\in ExB$ is $K_e$. The students enrolled in a class of the course $\lambda\in \Lambda$ are denoted as $S$. If a student $s\in S$ has answered an exercise $e\in ExB$, it is recorded as $<e, a_s>$ where $a_s$ is whether $s$'s answer is correct or not. 

\subsection{Student Score Estimation and Distribution}\label{sec:preest}
Given a question $q\in QuB$, the full score is $m(q)$. Suppose $m(q)$ is allocated to the knowledge points covered by the question (i.e., $K_q$). For one knowledge point $k\in K_q$, the score is denoted as $m(q,k)$ and $m(q)=\sum_{k\in K_q}m(q,k)$. The score of a student $s$ on question $q$ can be estimated as:
\begin{equation}
    m_s(q)=\sum_{k\in K_q} p_{s,k} m(q,k), 
\end{equation}
where $p_{s,k}$ is the mastery level of student $s$ on knowledge point $k$. The estimation of $p_{s,k}$ is based on \textit{Deep Knowledge Tracing} (DKT) \cite{piech2015DKT} and explained below. 

\begin{figure}[!t]
\centering
\includegraphics[scale=0.6]{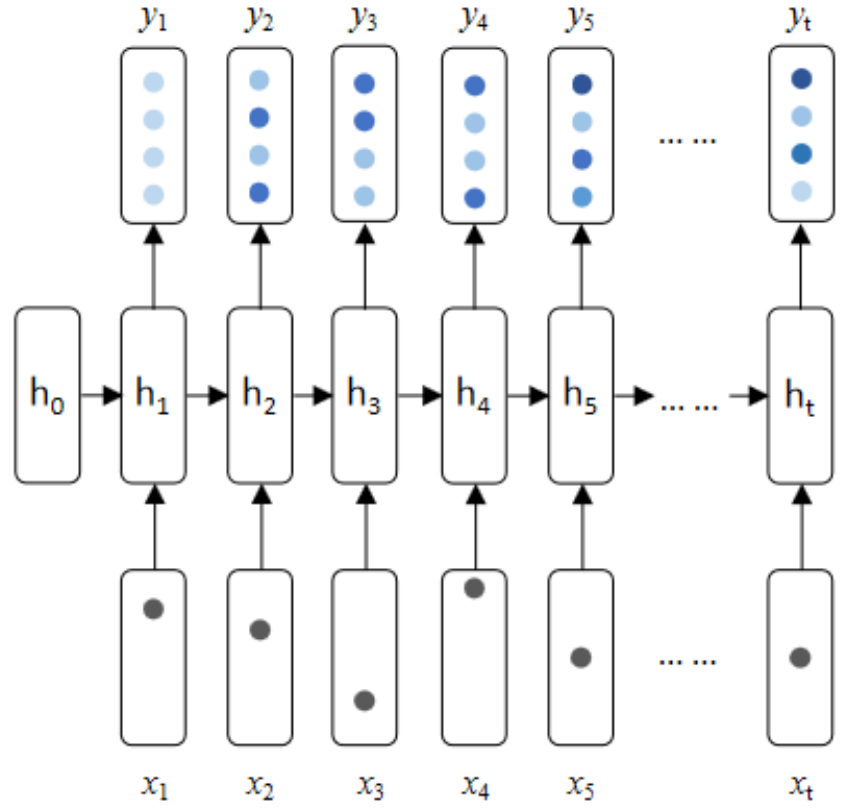}
\caption{Estimate the probability to correctly answer a knowledge point in a question with DKT.}
\label{fig:DKT}
\end{figure}

Knowledge tracing captures the knowledge mastery level of students according to the records of exercises answered so far by these students. In \cite{piech2015DKT}, \textit{Deep Knowledge Tracing} (DKT) model applies LSTM network to achieve this goal. The process is illustrated in Fig. \ref{fig:DKT}. The input $x_t$ is the encoded form of the knowledge point(s) covered by the exercise which have been answered by a student at time $t$, and the information whether the answer is correct or not; the output $y_t$ is a vector which comprise $|K|$ elements, each element corresponding to a unique knowledge point covered by the course. After training the model, the output of LSTM network can be used to predict the probability that a student $s$ would correctly answer a particular knowledge point based on the exercises done so far. Specifically, the value of $k$-th element in $y_t$ is $p_{s,k}$, i.e., the knowledge mastery level of student $s$ on knowledge point $k$.

Given an exam script $E$, the score estimation of student $s$ is the summation of scores that this student gets on all questions in $E$:
\begin{equation}\label{eq:score}
R_s(E) = \sum_{q\in E} m_s(q).
\end{equation}

Given a class of students $S$ and an exam script $E$, the expected score for each student can be obtained using Equ. (\ref{eq:score}). The distribution of students' expected scores is denoted as $\mathit{P}(S,E)$ which indicates the percentage of students at each score (e.g., $25\%$ at 70, $5\%$ at 90 if 100 is the full). 

Suppose $\mathit{P}^*$ is the desirable score distribution given by the teacher, that is, it is preferable that $\mathit{P}(S,E)$ is similar to $\mathit{P}^*$. In this study, without loss of generality, $\mathit{P}^*$ is a normal distribution where the mean $\mu$ represents the difficulty level of the exam script and the standard deviation $\sigma$ indicates the ability to distinguish academic performances between students. 

\subsection{Exam Script Generation Problem}
Given an integer $n$, a exam question bank $QuB$, and the desirable score distribution $\mathit{P}^*$, the exam script generation model $G(\Theta,n,S,QuB)$ outputs an exam script $E$ which consists of $n$ questions selected from $QuB$. Suppose the number of questions in $QuB$ is much greater than $n$, i.e., $|QuB|>>n$. The exam script generation is a modeling problem to optimize a set of parameters $\Theta$ so as to: 
\begin{itemize}
    \item minimize the difference between the probability of knowledge point $k$ to appear in $E$ and the relative importance of $k$ in the course (i.e., $\frac{w(k)}{\sum_{k'\in K} w(k')}$); and 
    \item minimize the difference between $\mathit{P}(S,E)$ and $\mathit{P}^*$. 
\end{itemize}
By minimizing the difference between $\mathit{P}(S,E)$ and $\mathit{P}^*$, the generated exam scripts are optimized to possess two important properties: (i) the expectation of student scores approaches to the mean of $\mathit{P}^*$ (i.e., the desirable difficulty level) and (ii) the standard deviation of student scores approaches to the standard deviation of $\mathit{P}^*$ (i.e., the desirable distinguishability on academic performances between students).
%I wish you the best of success.

 

\section{ExamGAN}\label{sec:model}
The proposed ExamGAN (Exam Script GAN) is a variant of conditional GAN model \cite{mirza2014cgan} \cite{goodfellow2014GAN}. Due to the remarkable effectiveness, GAN has been widely used in graphic image research, search optimization, and natural language processing (e.g., \cite{yu2017seqgan} \cite{lee2018RQGAN}). GAN comprises two major sub-models, i.e., \textit{generator} and \textit{discriminator}. In the model training, the generator continues generating fake data and feeds them to the discriminator as input. The discriminator needs learn to determine whether the input is from true data distribution or the fake data. Many variants of GAN have been proposed (e.g., \cite{pan2019GANsurvey} \cite{zamorski2019GANsurvey}). The conditional GAN is for the purpose of generating an output which is constrained by the input conditions \cite{mirza2014cgan}. It applies conditional settings in GAN such that both the generator and the discriminator take some kind of auxiliary information as input. 

\begin{figure}[ht]
	\centering
	\includegraphics[scale=0.4]{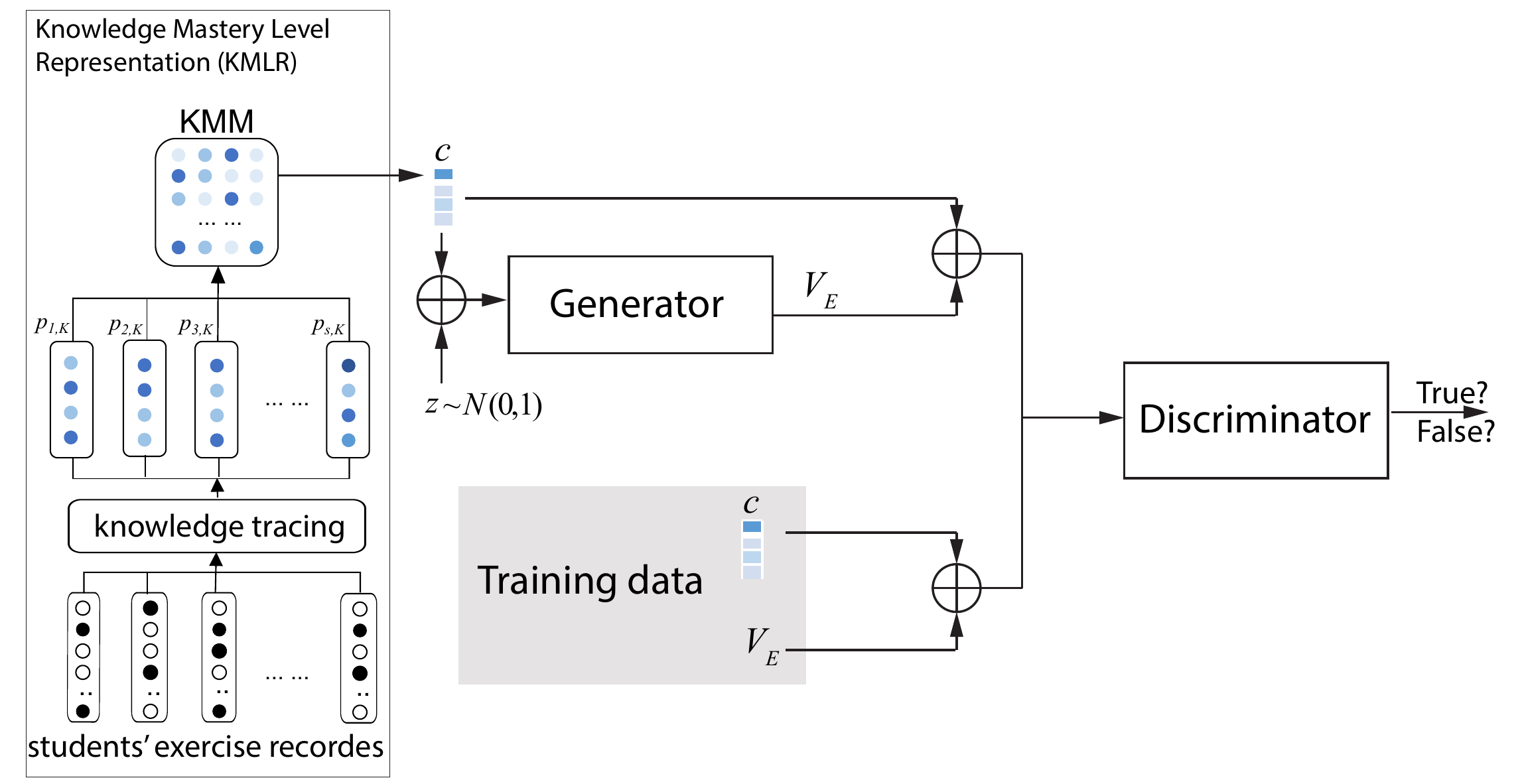}
	\caption{ExamGAN Model}
	\label{fig:ExamGAN}
 \end{figure}

As shown in Figure \ref{fig:ExamGAN}, ExamGAN mainly contains three components, i.e., \textit{knowledge mastery level representation} (KMLR), \textit{generator model} and \textit{discriminator model}. 

\subsection{Knowledge Mastery Level Representation (KMLR)}\label{sec:firstlayer}
As shown in Figure \ref{fig:ExamGAN}, the student exercise records are the input of KMLR. Using \textit{Deep Knowledge Tracing} \cite{piech2015DKT} (briefly introduced in Section \ref{sec:preest}), the knowledge mastery level of student $s$ on all knowledge points $K$ in this course is inferred, denoted as $p_{s,K}$. For all students, a knowledge mastery matrix (KMM) is obtained where each column corresponds to a student and each row corresponds to one of the $K$ knowledge points. The output of KMLR is denoted as $c$ which is derived from KMM. In $c$, each knowledge point keeps only two values, i.e., \textit{average} and \textit{standard deviation} of knowledge mastery levels of all students on the knowledge point.

\subsection{Generator and Discriminator Model}\label{sec:twolayers}
The generator model does not generate exam script $E$ directly. Instead, given the input condition $c$ from KMLR, it generates the probability distribution $V_E$ of all questions in the exam question bank to be selected in the exam script. The discriminator model checks whether the generated $V_E$ is desirable. The objective function of the generator and discriminator models is:
\begin{equation}\label{eq:examGAN} 
 \begin{split}
 \max_G\min_D V (D,G)=E_{x\sim p_{data}(x)}[\log D(x|c)]\\
 +E_{z\sim p_{z}(z)}[\log (1-D(G(z|c)))],
 \end{split}
\end{equation}
where $D$ and $G$ are discriminator model and generator model respectively; $x$ is the sample from the real data (the desirable exam script); $p_{data}(x)$ is the real data distribution; both generator and discriminator are conditioned on condition $c$ from KMLR; $z$ is a random variable sampled from a latent space with a prior noise distribution $P_z$. 

An ExamGAN is trained for course $\Lambda$. Given a collection of classes {$\lambda_1$, $\cdots$, $\lambda_i$} in $\Lambda$ where each class has a set of student exercise records, the training data for class $\lambda_i$ are the desirable exam scripts. Unfortunately, it is unrealistic to collect sufficient training data directly from teaching practices for a particular course. In this situation, we need an alternative approach to effectively obtain training data.  

\subsection{Desirable Properties of Exam Scripts}\label{sec:trainingdata}
Given a class of students, the high quality exam scripts should have desirable (i) \textit{validity} (i.e., the proper knowledge coverage), (ii) \textit{difficulty} (i.e., the expected average score of students), (iii) \textit{distinguishability} (i.e., the ability of distinguishing academic performances between students), and (iv) \textit{rationality} (i.e., the desirable student score distribution).

%(i) \textit{difficulty} (i.e., proper difficulty level), (ii) \textit{distinguishability} (i.e., the ability to distinguish academic performances between students, (iii) \textit{validity} (i.e., proper coverage of knowledge points), (iii) $\rationality$ (i.e., the desirable distribution of student score).   

%Given a class of students, we can quantitatively define the desirable property of exam script(s). In \cite{crocker1986test-theory,Yuan2012examination}, they have proposed the quality indicators of exam scripts, i.e., ,  and \textit{validity}. 

%The mean of all student's exam scores indicates the difficulty level of the exam script. It should be controlled near 70 if full mark is 100 \cite{LiuXP2003statistics}. The discrimination refers to the ability of the exam script to differentiate academic performances of students. According to \cite{Yuan2012examination}, the \textit{discrimination} is measured by $\frac{P_H-P_L}{100}$ where $P_H$ is the average of the $27\%$ the highest scores and $P_L$ is the average score of the $27\%$ lowest scores. The desirable value of discrimination is shown in Table \ref{tab:quality}. 

 \begin{table}
 \centering
	\caption{Quality of the exam script}
	\label{tab:quality}
	\begin{tabular}{cl}
		\hline
		distinguishability& Quality of the exam script\\
		\hline
		$ > 0.39 $& Excellent\\
		$ 0.30 - 0.39 $& Qualified\\
		$ 0.20 - 0.29 $& Passable and has possibility for improvement\\
		$ < 0.20 $& Should be discarded\\
		\hline
	\end{tabular}
  \end{table}

\subsubsection*{Desirable Difficulty and Distinguishability}\label{sec:variance}
The desirable \textit{difficulty} is that the average score of students is 70 if 100 is full \cite{LiuXP2003statistics}. According to \cite{Yuan2012examination}, the \textit{distinguishability} is measured by $\frac{P_H-P_L}{100}$ where $P_H$ is the average of the $27\%$ highest scores and $P_L$ is the average of the $27\%$ lowest scores. The desirable value of distinguishability is shown in Table \ref{tab:quality}, that is, $\frac{P_H-P_L}{100}>0.39$. 

\subsubsection*{Desirable Validity}
An exam script needs not cover all knowledge points. But the more important knowledge points should be more likely to be included in the exam scripts. Given a knowledge point $k\in K$, if $k$ is more important, it appears in more questions in the exam question bank of the course. That is, $w(k)$ is the frequency of $k$ in the exam question bank. The desirable knowledge point distribution in an exam script should be similar to the knowledge point distribution of the course. 

The knowledge point distribution of the course is denoted as $C_p$ whose value for each knowledge point is its frequency in the exam question bank; the knowledge point distribution of an exam script is denoted as $E_p$ whose value for each knowledge point is its frequency in the exam script. The cosine similarity is used to measure the similarity between $E_p$ and $C_p$. The higher similarity is preferable.  

\subsubsection*{Desirable Rationality}
Given an exam script of high quality, let $X$ be the variable of student score. $X$ should have a preferable distribution. Without loss of generality, we assume the preferable distribution is a normal distribution $\mathcal{Z}$ in a range of $0-100$ where the mean $\mu$, the average of student scores, should be 70 (i.e., the desirable difficulty level) and the standard deviation $\sigma$ should ensure excellent \textit{distinguishability} (Table \ref{tab:quality}). While it is straightforward $\mu=70$, we next derive the proper value of $\sigma$. Let $\alpha$ be the 27\%-quantile of $\mathcal{Z}$ and $\beta$ the 73\%-quantile of $\mathcal{Z}$. According to the definition of $Moment$ of normal distribution, The expectation of $X$ conditioned on the event that $X$ lies in an interval $[a,b]$ is given by
\begin{equation}
E[X|a \leq X \leq b] = \mu - \sigma^2\frac{f(b)-f(a)}{F(b)-F(a)},
\end{equation}
where $f$ and $F$ respectively are the density and the cumulative distribution function of $X$. We set $\overline{\alpha}$ represents the mean of the 27\% lowest values, and $\overline{\beta}$ represents the mean of the 27\% highest values. Then, $\overline{\alpha}$ can be expressed as:
\begin{equation}
\overline{\alpha} = \mu - \sigma^2\frac{f(\alpha)-f(0)}{F(\alpha)-F(0)},
\end{equation}
and $\overline{\beta}$ can be expressed as:
\begin{equation}
\overline{\beta} = \mu - \sigma^2\frac{f(100)-f(\beta)}{F(100)-F(\beta)},
\end{equation}
where $f(0) = 0$, $F(0) = 0$ and $f(100) = 0$, $F(100) = 1$. Furthermore, we set the distance between $\overline{\alpha}$ and $\overline{\beta}$ is $Dis$.
\begin{equation}
\begin{split}
Dis &= \overline{\beta} - \overline{\alpha} \\
    &= \sigma^2(\frac{f(\alpha)-f(0)}{F(\alpha)-F(0)}-\frac{f(100)-f(\beta)}{F(100)-F(\beta)})\\
    &= \sigma^2(\frac{f(\alpha)}{F(\alpha)} + \frac{f(\beta)}{1-F(\beta)}).\\
\end{split}
\end{equation}

\begin{figure}[ht]
\centering
\includegraphics[scale=0.6]{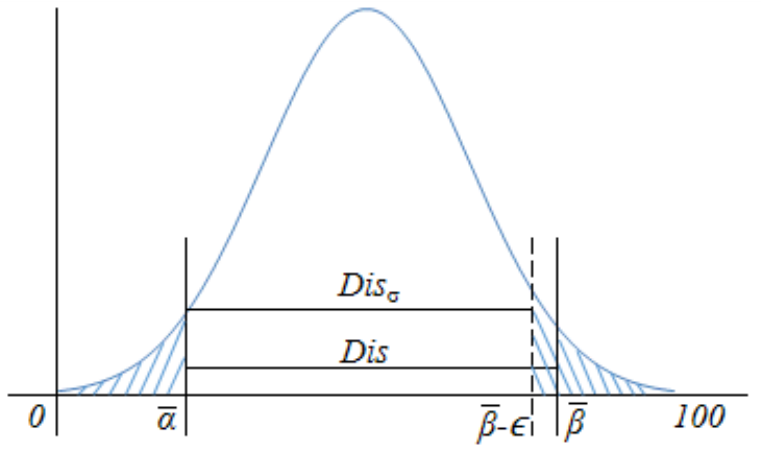}
\caption{Distance of two mean values.}
\label{fig:sigma}
\end{figure}

As shown in Figure \ref{fig:sigma}, we assume there is a point $\overline{\beta}-\epsilon$ on the $x$-axis which makes the areas of two cumulative distribution $F(\alpha)$ and $1-F(\beta-\epsilon)$ equal. $Dis_{\sigma}$ is the distance between $\overline{\alpha}$ and $\overline{\beta}-\epsilon$. Obviously, we have 
\begin{equation}
\begin{split}
Dis & \geq Dis_{\sigma}\\
    &= \sigma^2(\frac{f(\alpha)}{F(\alpha)} + \frac{f(\beta - \epsilon)}{1-F(\beta - \epsilon)})\\
    &= \sigma^2(\frac{f(\alpha)+f(\beta - \epsilon)}{F(\alpha)})\\
    &\geq \sigma^2(\frac{f(\alpha)}{F(\alpha)}),
\end{split}
\end{equation}
where $\alpha$ is the 27\%-quantile of $\mathcal{Z}$, then we can get $\alpha$ by checking the standard normal distribution table and do the normal distribution transformation, i.e,
\begin{equation}
\alpha = (-1.103063\sigma)+\mu,
\end{equation}
where we set $\mu = 70$, and let 
\begin{equation}\label{equ:sigma}
\sigma^2(\frac{f(\alpha)}{F(\alpha)}) = 39. 
\end{equation}

By substituting $\alpha$ into Equ. \eqref{equ:sigma}, we obtain the value of $\sigma$ which makes the distance between the average of the 27\% highest score and the average of the 27\% lowest scores greater than 39. That is, $\sigma=15$ leads to excellent distinguishability. 

Now, we have the mean and standard deviation of the desirable normal distribution $\mathcal{Z}$. Given an exam script $E$ for a class of students $S$, the student score distribution, denoted as $R_E^S$, should be similar to $\mathcal{Z}$. We use the Kullback-Leibler (KL) divergence \cite{James2011Kullback} as the similarity measure, that is, 
\begin{equation}\label{eq:dissim}
Dissim(R_E^S,\mathcal{Z}) = 1 - KL(P(R_E^S),P(Z)).  
\end{equation}

The higher similarity is preferable. 

\subsection{Creating Training Data}\label{sec:brutal}
Given a class of a course and the exercise records of students in the class, we randomly generate a large number of exam scripts from the exam question bank. Among them, those with desirable properties (discussed in Section \ref{sec:trainingdata}) are selected as the training data. This process is time consuming but it is acceptable to create training data offline. Once the generative model is trained with the training data, exam scripts can be online generated for any classes of the course. Given a class of students, the training data is created by taking the following steps:

\begin{description}
\item[S1.] The knowledge mastery level of students in the class, denoted as $c$, is generated based on the exercise records as explained in Section \ref{sec:firstlayer}. $c$ is the condition for the corresponding exam script.
\item[S2.] Among all knowledge points, we randomly pick up $n$ knowledge points based on their frequencies in the exam question bank, i.e., the more frequent one is more likely to be selected. 
\item[S3.] For each of the $n$ knowledge points, it may be covered by a number of questions from which one and only one is selected in random and inserted into the exam script. This is performed repeatedly until the exam script contains $n$ questions. By this way, $m$ ($m=1000$ in this work) exam scripts are created. 
\item[S4.] For each of the newly created $m$ exam scripts, denoted as $E_i$, the student score is estimated using knowledge tracing where it requires the records of exercises done by students as explained in Section \ref{sec:preest}. The distribution of estimated student scores $R_{E_i}^S$ is obtained; and the similarity between $R_{E_i}$ and $\mathcal{Z}$ is calculated following Equ. (\ref{eq:dissim}).
\item[S5.] Among all $m$ exam scripts, the $1\%$ with the highest similarity to $\mathcal{Z}$ are adopted. For each adopted exam script, an instance of training data $<c,V_E>$ is created where $V_E$ is the vector, each element corresponding to a question in the exam question bank. The elements in $V_E$ corresponding to the $n$ questions in the exam script are 1; and the value of other elements is 0.
\end{description}

To have more training data, we use all students who have exercise record in a course and form them into classes randomly. For each class, the above S1-S5 steps are repeated. 

\begin{algorithm}[ht]
    %\scriptsize %\small, \footnotesize, \scriptsize, or \tiny
	\caption{Training ExamGAN}\label{alg:examgan}
	\SetKwInOut{KIN}{Input}
	\SetKwInOut{KOUT}{Output}
	\KIN{Training data $<c,V_E>$}
	\KOUT{$G$}
	Initialize $\theta_{G}$ and $\theta_{D}$; \\
	\While{$\theta_{G}$ not converged} {
        \While{$\theta_{D}$ not converged} { 
            \tcp{Given a batch of training data $<c_i, V_{Ei}>$}
            \tcp{using $G$ to generate a batch of $<c_j, V_{Ej}>$}
	        $\theta_{D} \leftarrow \theta_{D} + \eta \nabla V_{D}(\theta_{D})$;
        }	    
	    $\theta_{G} \leftarrow \theta_{G} + \eta \nabla V_{G}(\theta_{G})$;
    }
\end{algorithm}

\subsection{Training and Using ExamGAN Model}
The training algorithm of ExamGAN is presented in Algorithm \ref{alg:examgan}. $<c_i,V_{E_i}>$ is an instance in the training data, i.e., a sample from $p_{data}(x)$ as shown in Equ. (\ref{eq:examGAN}). For $<c_j,V_{E_j}>$, $V_{E_j}$ is the output of generator on condition $c_j$. The discriminator model is trained to recognize the difference between $<c_i,V_{E_i}>$ and $<c_j,V_{E_j}>$; and the generator model is trained to generate $<c_j,V_{E_j}>$ such that the discriminator model cannot recognize their difference.   

Once ExamGAN has been trained, given any new class $\lambda_{new}$ in the course, the knowledge mastery level of students is identified following Section \ref{sec:firstlayer} and fed to the generator model. The output is the probability distribution of questions in the exam question bank to appear in the exam script. In the exam question bank, $n$ questions with the highest probabilities will be selected to form the new exam script for $\lambda_{new}$. 

\section{T-ExamGAN model}\label{sec:model2}
We extend ExamGAN to T-ExamGAN which generates a pair of equivalent exam scripts for the same class of students. The two exam scripts should have significantly different sets of questions while both of them retain the high quality. 

\begin{figure}[ht]
	\centering
	\includegraphics[scale=0.4]{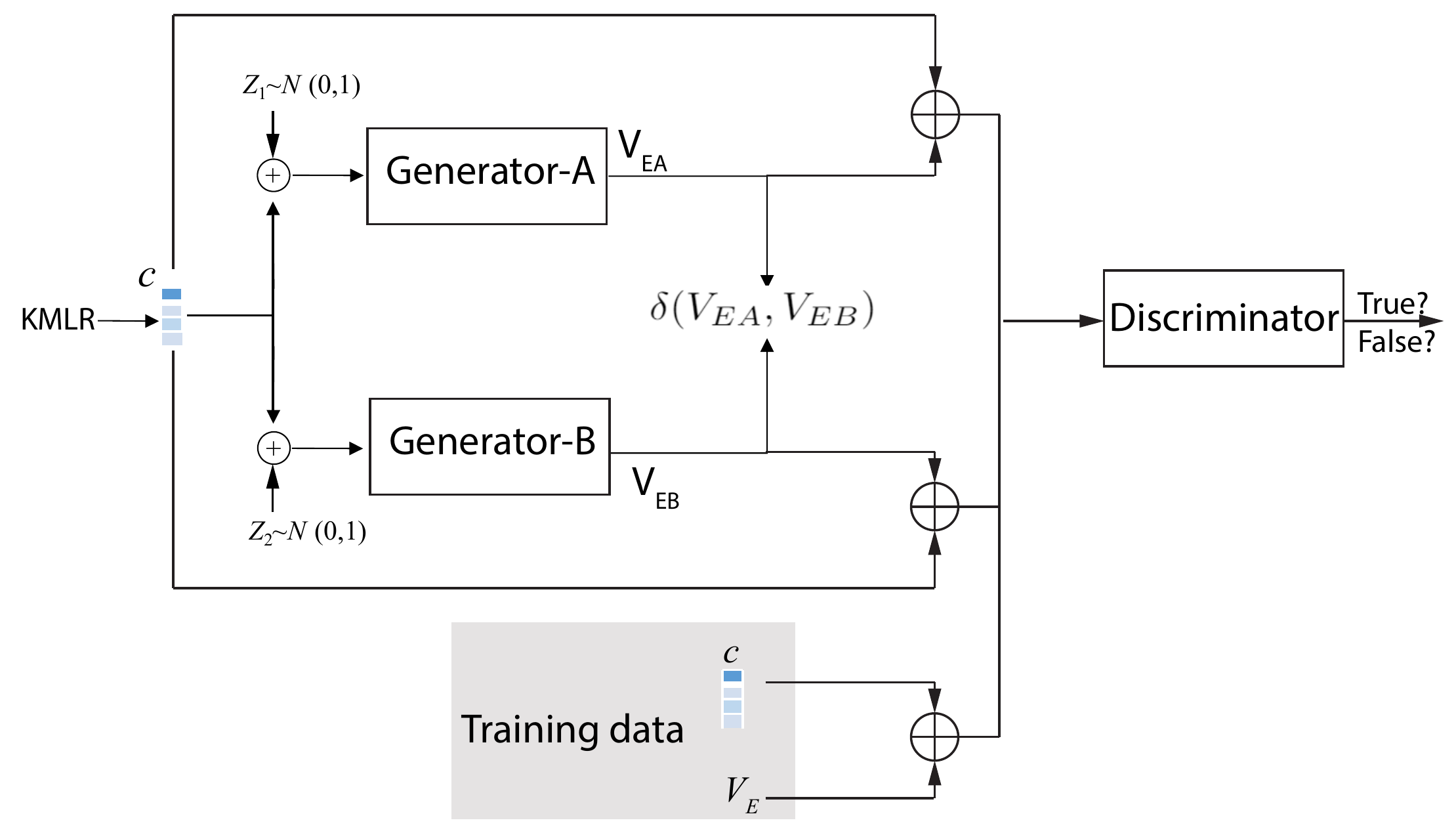}
	\caption{T-ExamGAN model.}
	\label{fig:TwinGAN}
\end{figure}

For a pair of exam scripts, they may have same questions but the percentage must be restrained. The Jaccard coefficient is applied to measure to which level two exam scripts have the same questions:
\begin{equation}\label{equ:sim}
f(E_A,E_B) = \frac{|E_A \cup E_B|-|E_A \cap E_B|}{|E_A \cup E_B|},
\end{equation}
where $E_A$ and $E_B$ represent a pair of exam scripts generated.  
 
\subsection{Framework of T-ExamGAN}
Difference from ExamGAN, T-ExamGAN comprises four components, i.e., \textit{knowledge mastery level representation} (KMLR), \textit{generator-A}, \textit{generator-B} and \textit{discriminator}, as shown in Figure \ref{fig:TwinGAN}. 

KMLR is exactly same as that in ExamGAN. As introduced in Section \ref{sec:firstlayer}, the output of KMLR is the condition $c$ of generators. The generator-A (denoted as $G_A$) outputs $V_{E_A}$ and the generator-B (denoted as $G_B$) outputs $V_{E_B}$. $V_{E_A}$ ($V_{E_B}$) is the probability distribution of all questions in the exam question bank to appear in the exam script $E_A$ ($E_B$). To ensure $V_{E_A}$ and $V_{E_B}$ have significantly different sets of questions, $G_A$ and $G_B$ are trained alternatively. The loss function is cross entropy between $V_{E_A}$ and $V_{E_B}$.
\begin{equation}
H(V_{E_A},V_{E_B})=-\sum _{x\in QuB}V_{E_A}(x)\,\log V_{E_B}(x), 
\end{equation}
where $V_{E_*}(x)$ is the probability of question $x$ in $V_{E_*}$. If the loss function $H(V_{E_A},V_{E_B})$ is greater, $V_{E_A}$ and $V_{E_B}$ are more similar and thus one of the generators needs to be adjusted so as to reduce the loss function value (i.e., making $V_{E_A}$ and $V_{E_B}$ less similar). 

At the same time, to ensure the high quality of the generated exam scripts by $G_A$ and $G_B$, the discriminator model checks $V_{E_A}$ and $V_{E_B}$; and the generator $G_A$ and $G_B$ are optimized accordingly as in ExamGAN. Note the training data are also same as those in ExamGAN. 

\subsection{Training T-ExamGAN}
T-ExamGAN is defined as the following minmax game.
 \begin{equation}
 \begin{split}
 \mathop{min}\limits_{G_A,G_B}\mathop{max}\limits_{D}V(G_A,G_B,D)= & E_{E\sim p_{data}(E)}[log(D(E|c))]\\ 
  +E_{z\sim p_{z}(z)}&[1-log(D(G_A(z|c)))]\\  
  +E_{z\sim p_{z}(z)}&[1-log(D(G_B(z|c)))]\\ 
  +\lambda \vert \psi-&H(V_{GA},V_{GB}) \vert.
 \end{split}
 \end{equation}
 
The objective function of discriminator can be expressed as:
 \begin{equation}
 \begin{split}
 V_D=\frac{1}{r} \sum_{i=1}^{r}{log(D(E|c))}+ \frac{1}{r} \sum_{i=1}^{r}{(1-log(D(G_A(z|c))))}\\
 +\frac{1}{r} \sum_{i=1}^{r}{(1-log(D(G_B(z|c))))}.
 \end{split}
 \end{equation}
 
 We maximize the objective function by updating discriminator parameters $\theta_{D}$:
 \begin{equation}
 \theta_{D} \leftarrow \theta_{D} + \eta \nabla V_{D}(\theta_{D}),
 \end{equation} 
 where $\eta$ is the parameter of step size and so in the following equations.

For training generator $G_A$ and $G_B$, the objective functions are below respectively:
 \begin{equation}V_{G_A}=\frac{1}{r} \sum_{i=1}^{r}{log(D(G_A(z|c)))},
 \end{equation}
 \begin{equation}V_{G_B}=\frac{1}{r} \sum_{i=1}^{r}{log(D(G_B(z|c)))},
 \end{equation}
where $r$ is the number of the training sample. We maximize the objective function by updating parameters $\theta_{G_A}$ and $\theta_{G_B}$:
 \begin{equation}
 \theta_{G_A} \leftarrow \theta_{G_A} + \eta \nabla V_{G_A}(\theta_{G_A}).
 \end{equation}
 
 \begin{equation}\theta_{G_B} \leftarrow \theta_{G_B} + \eta \nabla V_{G_B}(\theta_{G_B}).
 \end{equation}
 
To make the exam scripts generated by generator $G_A$ and generator $G_B$ with different sets of questions, we aim to minimize 
\begin{equation}
    \mathcal{L}(c) = \vert \psi-H(V_{E_A},V_{E_B}) \vert,
\end{equation}
where $\psi$ is a hyper-parameter indicating the required difference between the $E_A$ and $E_B$. The objective function $\mathcal{L}(c)$ is optimized by updating generator parameters $\theta_{G_B}$ and $\theta_{G_A}$:
\begin{equation}
\theta_{G_A} \leftarrow \theta_{G_A} + \eta \nabla \mathcal{L}(c)(\theta_{G_A}). 
\end{equation}

\begin{equation}
\theta_{G_B} \leftarrow \theta_{G_B} + \eta \nabla \mathcal{L}(c)(\theta_{G_B}).
\end{equation}

Note it is less intuitive to specify the hyper-parameter $\psi$. So, we take an alternative approach. $E_A$ ($E_B$) is formed by the $n$ questions with the highest probabilities in $V_{E_A}$ ($V_{E_B}$). According to Equ. (\ref{equ:sim}), if $f(E_A,E_B)$ is no more than a specified ratio, say $30\%$, the stop criteria of optimization is met.

\subsection{Two Training Strategies}
Two training strategies can be applied, i.e., \textit{individual quality priority} and \textit{twin difference priority}.  

\subsubsection{Individual Quality Priority}
The optimization priority is on the quality of individual exam script. Algorithm~\ref{alg:one} illustrates the training process. (1) The generators and discriminator are trained simultaneously. The goal is that the discriminator should be able to identify true and false samples, and each generator generates fake samples to deceive the discriminator. Since the generators are trained separately, for each of them it is same as training ExamGAN. This step is repeated until two generators have been well trained. (2) The difference between the generated exam scripts from two generators are checked. The parameters in one of the two generators will be updated once. The above two steps are repeated by $\gamma$ times.  

 \begin{algorithm}[ht]
    %\scriptsize %\small, \footnotesize, \scriptsize, or \tiny
	\small
	\SetKwInOut{KIN}{Input}
	\SetKwInOut{KOUT}{Output}
	\caption{T-ExamGAN Training - Individual Quality Priority}\label{alg:one}
	\KIN{Training data $<c,V_E>$; $\gamma$}
	\KOUT{$G_{A},G_{B}$}
	Initialize $\theta_{G_A}$, $\theta_{G_B}$, $\theta_{D}$, $r=0$; \\
	\While{$r<\gamma$} {
	    $r=r+1$; \\
	    \While{$\theta_{G_A}$, $\theta_{G_B}$ not converged} {
	        \While{$\theta_{D}$ not converged} {
                \tcp{Given training data $<c_i, V_{Ei}>$s} 
                \tcp{using $G_A$ generate $<c_j, V_{Ej}>$s} 
                \tcp{using $G_B$ generate $<c_l, V_{El}>$s} 
	            $\theta_{D} \leftarrow \theta_{D} + \eta \nabla V_{D}(\theta_{D})$; 
            }
	        $\theta_{G_A} \leftarrow \theta_{G_A} + \eta \nabla V_{G_A}(\theta_{G_A})$;	\\
	        $\theta_{G_B} \leftarrow \theta_{G_B} + \eta \nabla V_{G_B}(\theta_{G_B})$;	
        }
        \If{$r$ is even}{
            $\theta_{G_A} \leftarrow \theta_{G_A} + \lambda \nabla \mathcal{L}(c)(\theta_{G_A})$; \\
        }\Else{
            $\theta_{G_B} \leftarrow \theta_{G_B} + \lambda \nabla \mathcal{L}(c)(\theta_{G_B})$; 
        }
    }
 \end{algorithm}

\subsubsection{Twin Difference Priority}
It takes the difference between the generated exam scripts as the primary goal. Algorithm~\ref{alg:two} illustrates the training process. (1) The two generators are trained if the difference between the exam scripts does not meet the requirement, i.e., the parameters in two generators are alternatively updated to increase the difference. The training stops until the difference requirement is met. (2) The generators and discriminator are trained simultaneously. The goal is that the discriminator should be able to identify true and false samples, and each generator generates fake samples to deceive the discriminator. The generators are trained separately, for each of them it is same as training ExamGAN. Note that the parameters of discriminator and generators are updated once only. The above two steps are repeated by $\gamma$ times.  

\begin{algorithm}[ht]
    %\scriptsize %\small, \footnotesize, \scriptsize, or \tiny
    \small
	\SetKwInOut{KIN}{Input}
	\SetKwInOut{KOUT}{Output}
	\caption{T-ExamGAN Training - Difference Priority}\label{alg:two}
	\KIN{Training data $<c,V_E>$; $\gamma$}
	\KOUT{$G_{A},G_{B}$}
	Initialize $\theta_{G_A}$, $\theta_{G_B}$, $\theta_{D}$, $r=0$; \\
	\While{$r<\gamma$} {
        $r=r+1$; \\	
        $l = 0$; \\
	    \While{$\theta_{G_A}$, $\theta_{G_B}$ not converged} {
            \tcp{using $G_A$ generate $<c_j, V_{Ej}>$s} 
            \tcp{using $G_B$ generate $<c_l, V_{El}>$s} 
            \If{$l$ is even}{
                $\theta_{G_A} \leftarrow \theta_{G_A} + \lambda \nabla \mathcal{L}(c)(\theta_{G_A})$; 
            }\Else{
                $\theta_{G_B} \leftarrow \theta_{G_B} + \lambda \nabla \mathcal{L}(c)(\theta_{G_B})$; 
            }
            $l=l+1$;
        }
        \While{$\theta_{D}$ not converged} {
            \tcp{Given training data $<c_i, V_{Ei}>$s} 
            \tcp{using $G_A$ generate $<c_j, V_{Ej}>$s} 
            \tcp{using $G_B$ generate $<c_l, V_{El}>$s} 
	        $\theta_{D} \leftarrow \theta_{D} + \eta \nabla V_{D}(\theta_{D})$;
        }
	    $\theta_{G_A} \leftarrow \theta_{G_A} + \eta \nabla V_{G_A}(\theta_{G_A})$;	\\
	    $\theta_{G_B} \leftarrow \theta_{G_B} + \eta \nabla V_{G_B}(\theta_{G_B})$;	
    }
\end{algorithm}

\section{Experiments}\label{sec:test}
The experiments evaluate the performance of proposed ExamGAN and T-ExamGAN. All tests were run on a desktop computer and a laptop. The desktop computer with Intel(R) Core i9 Duo 2.4GHz CPU, and 8GB of main memory. The laptop with Intel(R) Core i7 Duo 2.4GHz CPU, and 8GB of main memory. The code used for programming is python 3.7, and sklearn 0.20.3. The development environment for deep learning is Tensorflow 1.10.0 and Keras 2.2.4. 

All tests explore three datasets. Two datasets are from \textit{ASSISTments} \cite{google2014assistments} which is an electronic tutor teaching and evaluating students in grade-school math. One of the two datasets was gathered in the school year 2009-2010 (denoted as \textit{assistments0910}) and the other was gathered in 2012-2013 (denoted as \textit{assistments1213}). The third dataset named \textit{OLI Engineering Statics} is from a college-level engineering statics course (denoted as \textit{olies2011}); the dateset is available online (pslcdatashop.web.cmu.edu). The detailed information of the three datasets are shown in Table \ref{tab:Data set}. Note exercise questions in Table \ref{tab:Data set} are unnecessary to be the exam questions. To test different situations, 10,000 exercise questions for \textit{assistments0910} and \textit{assistments1213} are randomly selected to form the exam question bank; for \textit{olies2011}, the exam question bank includes the exercise questions and synthetic questions generated following the distribution of real questions on knowledge points.  

For each dataset, 100 classes are formed randomly, each with 50 students. For each class, 10 training exam scripts are generated following Section \ref{sec:brutal}. Based on records of exercises answered, the deep knowledge tracing (DKT) model is trained to estimate the knowledge mastery levels of students in each class. The training set includes 80 classes, the validation set 10 classes and the test set 10 classes.
 
 \begin{table}
 \centering
  \caption{Overview of Datasets (KPs: Knowledge Points, EQs: Exercise Questions)}
  \label{tab:Data set}
  \begin{tabular}{ccccl}
    \hline
    \textbf{Dataset} & \textbf{KPs} & \textbf{Students} & \textbf{EQs} & \textbf{Records}\\
    \hline
    \textbf{assistments0910} & 123& 4,163& 26,688& 278,607\\
    \textbf{assistments1213} & 265& 28,834& 53,091& 2,506,769\\
    \textbf{olies2011}    & 85& 335& 1,223& 45,002\\
  \hline
\end{tabular}
\end{table}

\subsection{Effectiveness of ExamGAN}\label{sec:effexamgan}
Given a class of students, the high quality exam scripts should have desirable (i) \textit{validity} (i.e., the proper knowledge coverage), (ii) \textit{difficulty} (i.e., the expected average score of students), (iii) \textit{distinguishability} (i.e., the ability of distinguishing academic performances between students), and (iv) \textit{rationality} (i.e., the desirable student score distribution). 

As discussed in Section \ref{sec:trainingdata}, the desirable \textit{validity} is that the knowledge point distribution in exam script is similar to that in exam question bank; the desirable \textit{difficulty} is that the students' average score is close to $70$ (the full score $100$); the desirable \textit{distinguishability} is that the difference between average of the $27$\% highest score and the average of the $27$\% lowest scores is greater than $0.39$ (as shown in Table \ref{tab:quality}); the desirable \textit{rationality} is that the student score distribution is similar to a user specified distribution (i.e., normal distribution in range $0$-$100$ with mean $70$ and standard deviation $15$ in this work).  

\subsubsection{Baseline Methods}
We compare ExamGAN against three baseline methods which represent the state-of-the-art in automatic exam script generation. 
\begin{itemize}
\item \textit{Random Sampling and Filtering} (RSF) randomly samples a subset of questions from the exam question bank to form an exam script as detailed in Section \ref{sec:brutal}.
\item \textit{Genetic Algorithm} (GA) regards an exam script as an individual and each question is regarded as a chromosome. The expected difficulty level and the coverage of knowledge points are taken as the goal of evolution \cite{yildirim2010Genetictpg}. In GA, the unique parameters of genetic operation include the \textit{crossover rate} ($0.8$ by default), the \textit{mutation rate} ($0.003$ by default), and the size of population ($1000$ by default). 
\item \textit{Conditional Variational Autoencoder} (ExamVAE) is a VAE (briefed in Section \ref{sec:related}) conditioned on $c$ from KMLR as ExamGAN. Also, both ExamGAN and ExamVAE assume the exam scripts are generated by the same prior latent distribution $z$, i.e., normal distribution $N(0,1)$ \cite{doersch2016tutorofVAE}.
\end{itemize}

\begin{figure*}[ht]
	\centering
	\includegraphics[scale=0.57]{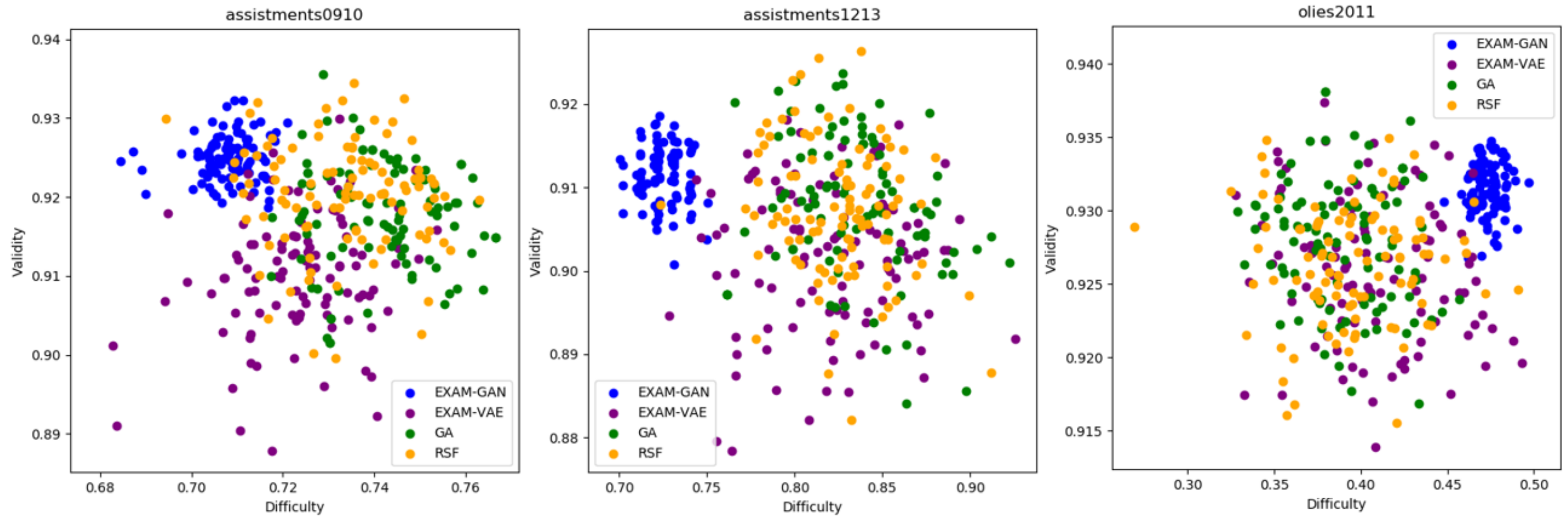}
	\caption{Validity vs. Difficulty.}
	\label{fig:dif-valid}
 \end{figure*}
 \begin{figure*}[ht]
	\centering
	\includegraphics[scale=0.57]{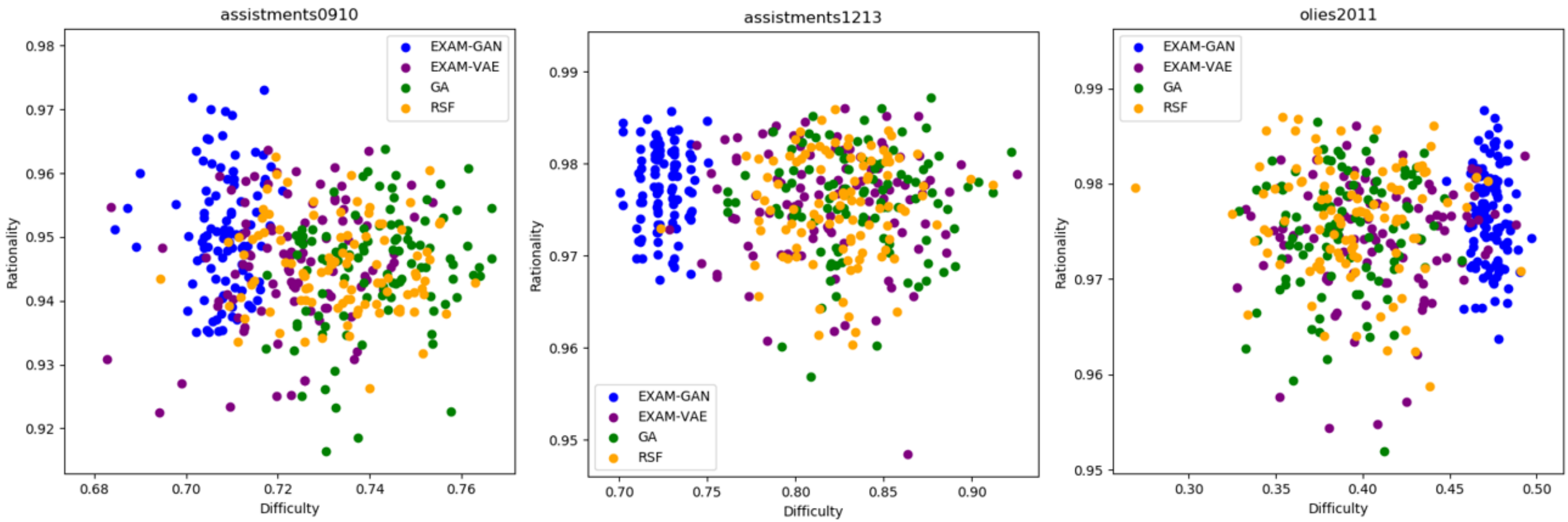}
	\caption{Rationality vs. Difficulty.}
	\label{fig:dif-Ration}
 \end{figure*}
 \begin{figure*}[ht]
	\centering
	\includegraphics[scale=0.57]{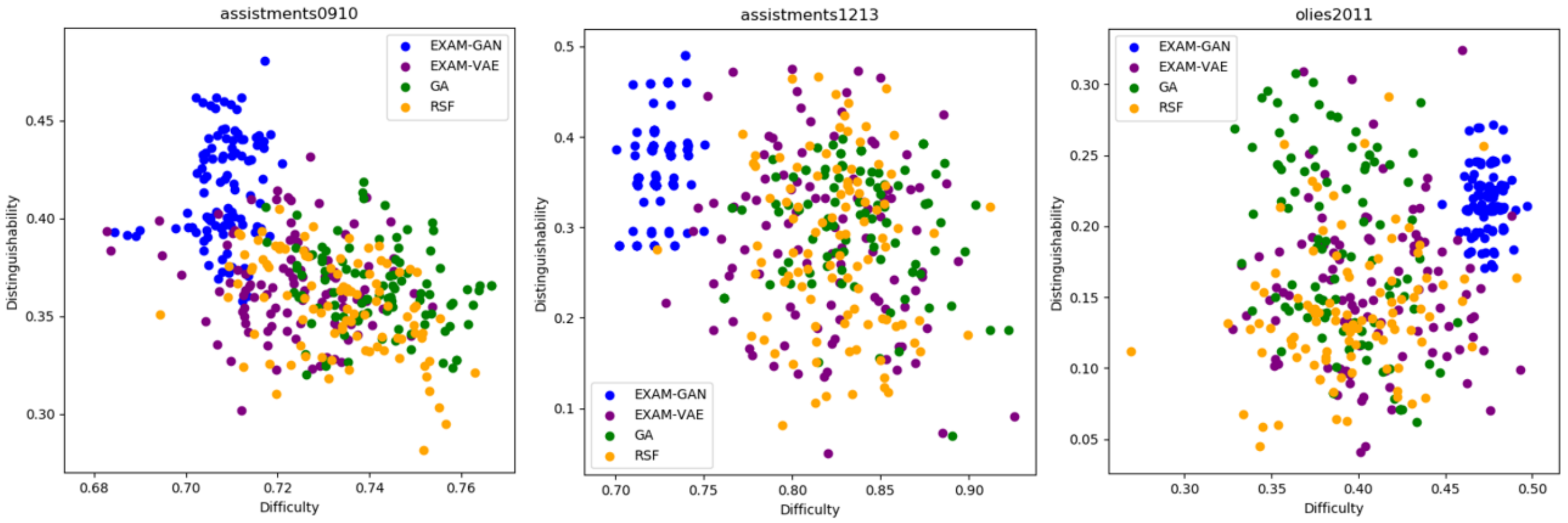}
	\caption{Distinguishability vs. Difficulty.}
	\label{fig:dif-Distin}
 \end{figure*}

While RSF and GA do not require training datasets, ExamVAE uses the training dataset as ExamGAN and T-ExamGAN. In ExamGAN, the generator is a two-layer network where the first layer uses the \textit{sigmoid} activation function and the second layer using the \textit{tanh} activation function; the discriminator is also a two-layer network where both layers use the \textit{sigmoid} activation function. Model optimization is based on \textit{gradient descent optimizer} \cite{ruder2016gdoa} with a learning rate $0.001$. For hyper-parameter optimization, we set dropout rate to $0.3$ for all layers.

In ExamVAE, the encoder consists of three fully connected layers where the first is the input layer and the other two calculate the mean of the training samples and the mean of the corresponding input conditions. The decoder consists of a custom layer and two fully connected layers where the custom layer receives input noise and re-parameter changes \cite{doersch2016tutorofVAE} and the other two layers decode the output. The activation function of encoder is \textit{ReLU} and the activation function of decoder is \textit{ReLU} and \textit{sigmoid}. The model is trained by using \textit{Adam Optimizer} \cite{kingma2015adam} with a learning rate $0.01$.

\begin{figure*}[ht]
	\centering
	\includegraphics[scale=0.47]{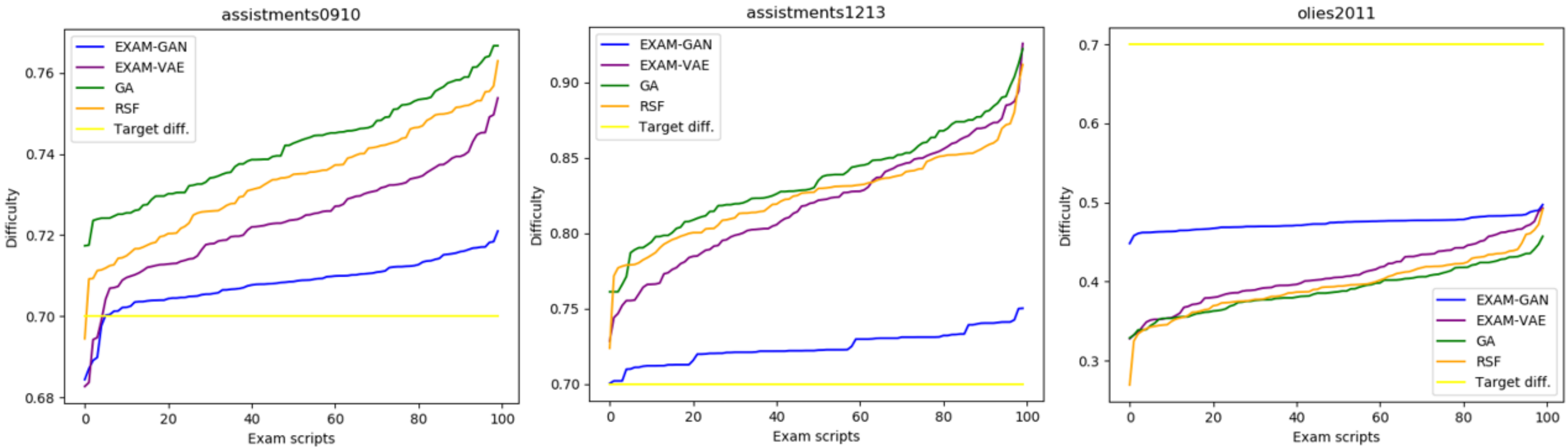}
	\caption{Comparison of Difficulty.}
	\label{fig:diff}
 \end{figure*}
 \begin{figure*}[ht]
	\centering
	\includegraphics[scale=0.47]{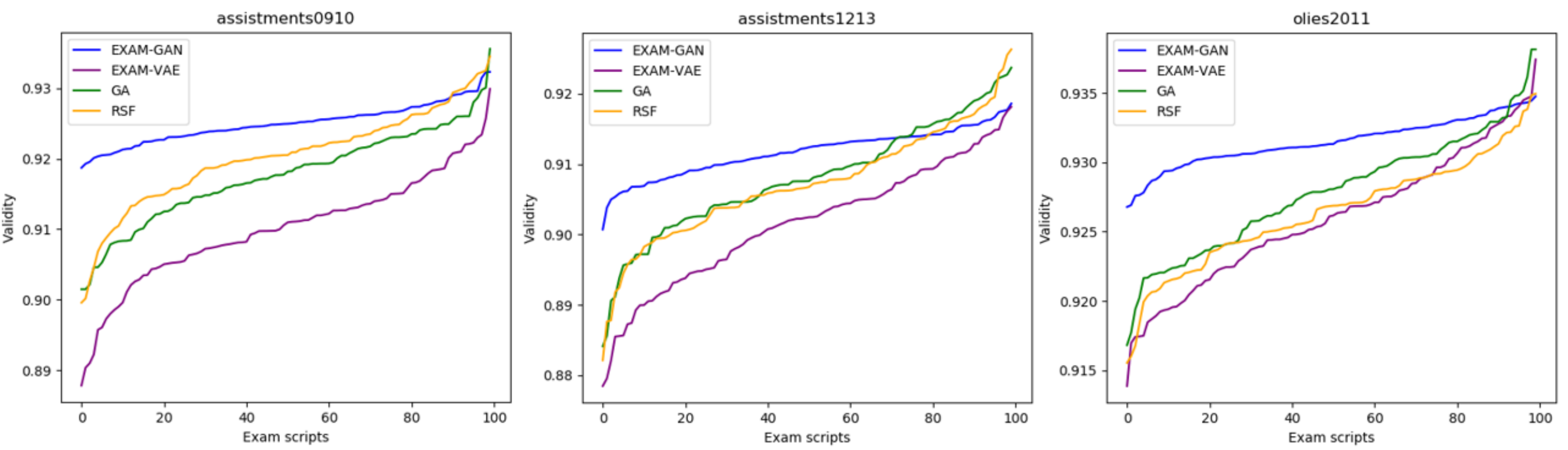}
	\caption{Comparison of Validity.}
	\label{fig:valid}
 \end{figure*}
 \begin{figure*}[ht]
	\centering
	\includegraphics[scale=0.47]{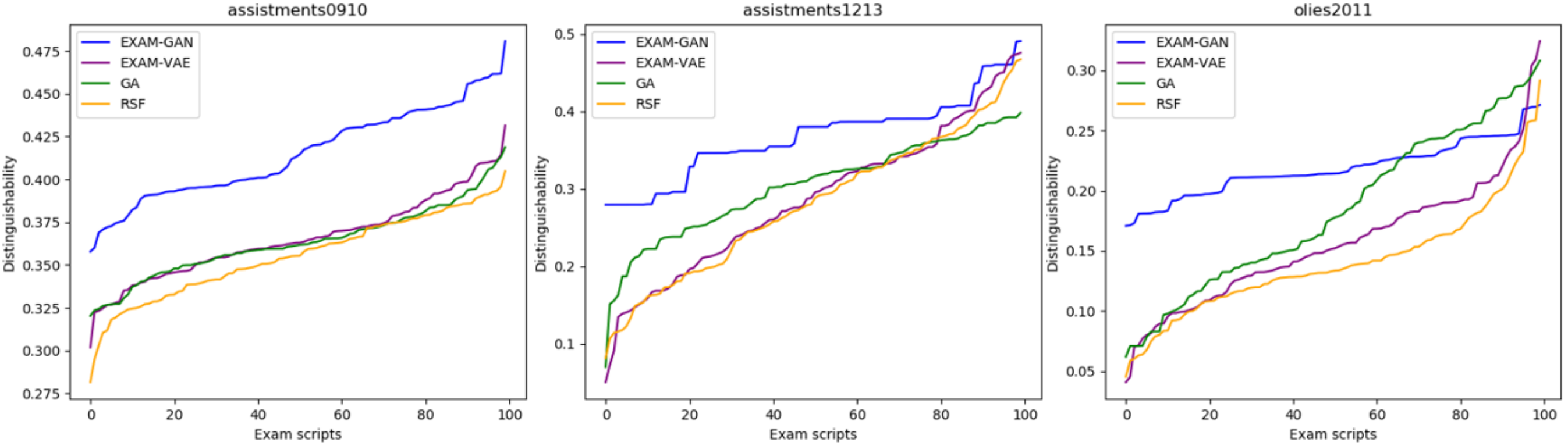}
	\caption{Comparison of Distinguishability.}
	\label{fig:distin}
 \end{figure*}
 \begin{figure*}[ht]
	\centering
	\includegraphics[scale=0.47]{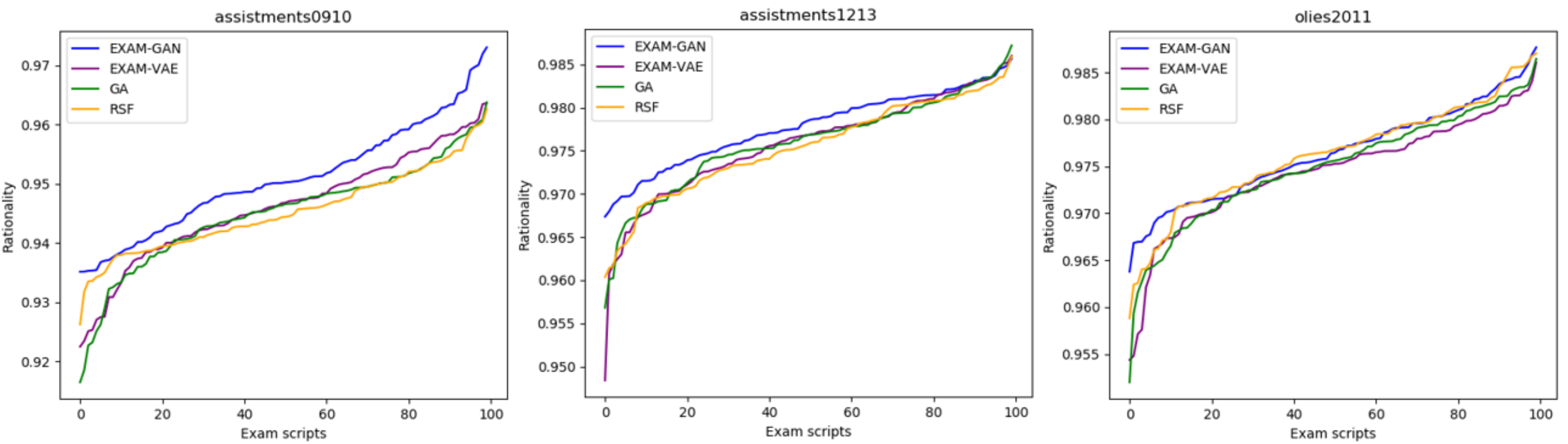}
	\caption{Comparison of Rationality.}
	\label{fig:ral}
 \end{figure*}

  \begin{figure*}[ht]
	\centering
	\includegraphics[scale=0.57]{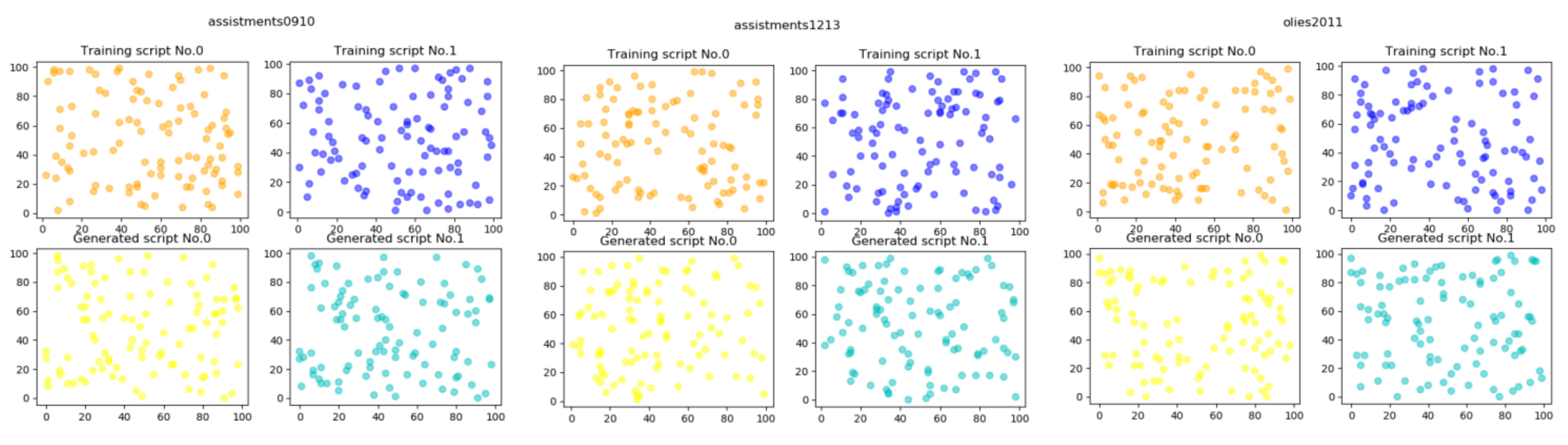}
	\caption{Questions in training exam scripts vs. questions in generated exam scripts.}
	\label{fig:16}
 \end{figure*}

\begin{figure}[ht]
	\centering
	\includegraphics[scale=0.28]{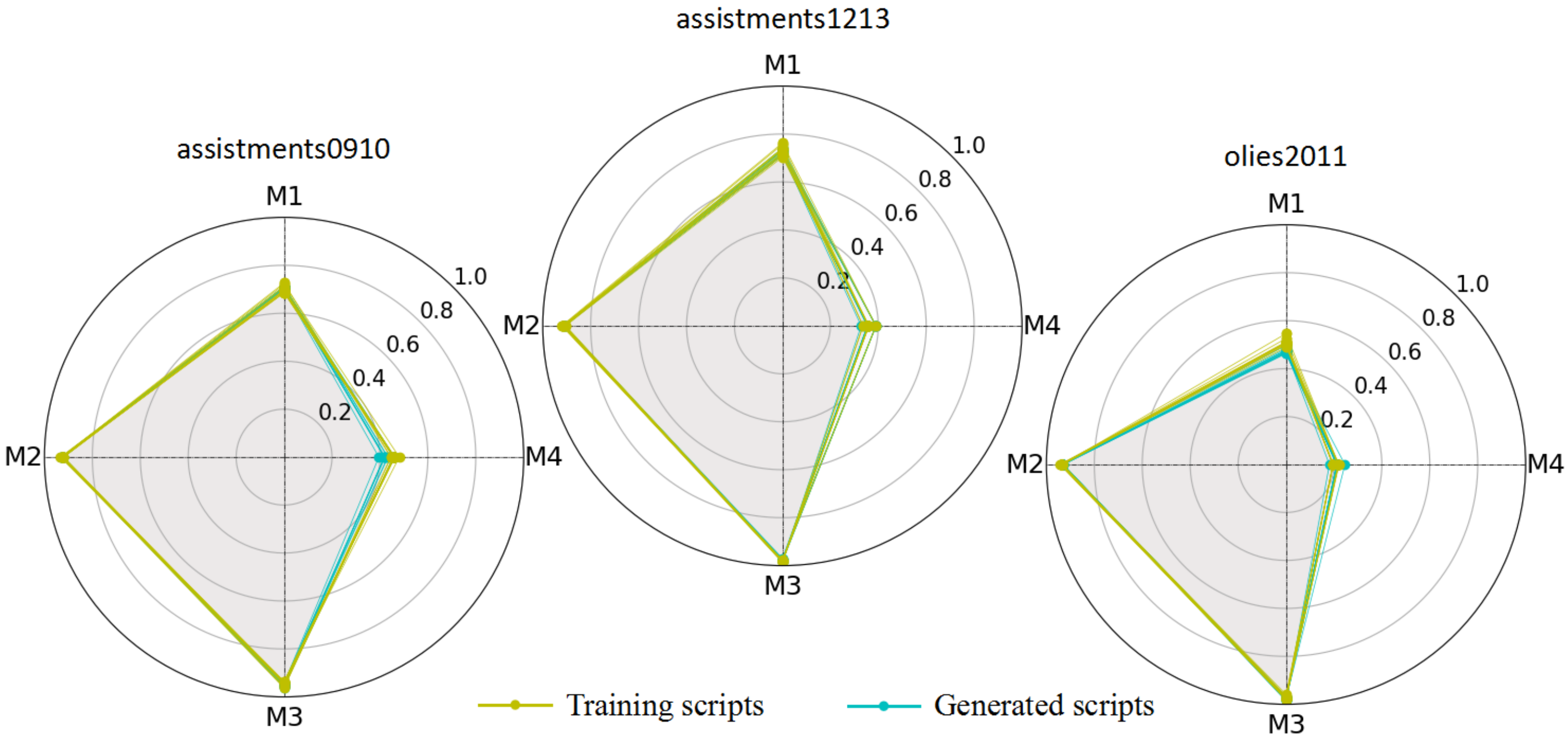}
	\caption{Training exam script vs. generated exam scripts in terms of the four properties (M1-\textit{Difficulty}, M2-\textit{Validity}, M3-\textit{Rationality}, M4-\textit{Distinguishability)}}
	\label{fig:15}
 \end{figure}

\subsubsection{Experiment Results}
Fig. \ref{fig:dif-valid}, \ref{fig:dif-Ration}, and \ref{fig:dif-Distin} present the quality of exam scripts generated using ExamGAN and baselines (each generates $100$) on the three datasets. Fig. \ref{fig:dif-valid} is \textit{validity} vs. \textit{difficulty}; Fig. \ref{fig:dif-Ration} is  \textit{rationality} vs. \textit{difficulty}; Fig. \ref{fig:dif-Distin} is \textit{distinguishability} vs. \textit{difficulty}. Compared with the baselines, the exam scripts generated using ExamGAN tend to be closer to the desirable \textit{difficulty} value ($0.7$), closer to the desirable \textit{validity} value (closer to 1 is better), closer to the desirable \textit{rationality} value (closer to 1 is better), closer to the desirable \textit{distinguishability} value ($>0.39$). Also, the quality of exam scripts generated using ExamGAN are more reliable in general in the four aspects since they are more concentrated. 

Interestingly, it is hard for dataset \textit{assistments1213} to generate exam scripts with difficulty level greater than $0.5$. It implies that the questions in the exam question bank are too easy considering the knowledge mastery level of the students in the class based on records of exercises answered. This reveals the side benefit of this study is to verify whether an update of the exam question bank is necessary to maintain the quality assessment.   

Fig. \ref{fig:diff}, \ref{fig:valid}, \ref{fig:distin}, and \ref{fig:ral} further illustrate the quality of generated exam scripts on the three datasets. In Fig. \ref{fig:diff}, for each method, the generated 100 exam scripts are sorted in ascending order in terms of \textit{difficulty} along $x$-axis. In Fig. \ref{fig:valid}, \ref{fig:distin}, and \ref{fig:ral}, the exam scripts are sorted in ascending order in terms of \textit{validity}, \textit{rationality} and \textit{distinguishability} along $x$-axis respectively. In Fig. \ref{fig:diff}, it clearly shows the exam scripts generated using ExamGAN are always closer to the desirable \textit{difficulty} (indicated by the yellow line). In Fig. \ref{fig:valid}, \ref{fig:distin}, and \ref{fig:ral}, the exam scripts generated using ExamGAN are obviously better than that using baselines in terms of \textit{validity} (closer to 1 is better), \textit{distinguishability} ($>0.39$), \textit{rationality} (closer to 1 is better) respectively in most cases. 

The conclusion is that the exam scripts generated using ExamGAN are more likely to have the desirable difficulty level, the better knowledge point coverage, the desirable student score distribution, and the better ability to distinguish academic performances between students. 

\subsubsection{Training Scripts vs. Generated Scripts}
Fig. \ref{fig:15} illustrates the properties of 15 randomly selected training exam scripts of 5 classes (3 for each) from the test set, and these of 2 generated exam scripts of the same 5 classes. Clearly, the generated exam scripts have the highly similar properties as the training exam scripts. Fig. \ref{fig:16} illustrates that the generated exam scripts include different sets of questions from the training exam scripts. In the test set of each dataset, one class is randomly selected and the questions of two training exam scripts are highlighted in the exam question bank (Training Script 0 and 1). For the same class, the questions of two generated exam scripts using ExamGAN are also highlighted in the exam question bank (Generated Script 0 and 1).

\subsection{Effectiveness of T-ExamGAN}
Using T-ExamGAN, a pair of exam scripts are generated. In the experiments, we compare (i) the two training strategies of T-ExamGAN, and (ii) T-ExamGAN against a baseline based on ExamGAN where two exam scripts are generated independently using the ExamGAN twice to form a pair. Here, the ExamGAN-based baseline is implemented in two settings: the first is denoted as ExamGAN@500 where ExamGAN is trained by $500$ epochs; the second is denoted as ExamGAN@1500 where ExamGAN is trained by $1500$ epochs. 
 
\subsubsection{Training T-ExamGAN} 
As introduced in Section \ref{sec:model2}, T-ExamGAN has two generators. The two generators are initialized by training each individually as training ExamGAN. Both are trained by 1500 epochs. The discriminator in T-ExamGAN is same as the discriminator in ExamGAN. The training datasets for T-ExamGAN are same as those for training ExamGAN. After initializing the two ExamGAN models, T-ExamGAN is trained. Using \textit{individual quality priority} training strategy, the T-ExamGAN model is denoted as T-ExamGAN@S1; using \textit{twin difference priority} training strategy 2, the T-ExamGAN model is denoted by T-ExamGAN@S2.  

\subsubsection{Experiment Results}
For T-ExamGAN@S1, T-ExamGAN@S2, ExamGAN@500 and ExamGAN@1500, each generates 100 pairs of exam scripts. It is desirable that (i) both two exam scripts have the similar quality in terms of \textit{difficulty}, \textit{distinguishability}, \textit{validity} and \textit{rationality}; (ii) the two exam scripts have significantly different sets of questions. 

%Among the 100 pairs of exam scripts generated using each method, the difference between the two exam scripts in each pair is measured in terms of quality (in this study, we use the Euclidean distance of four dimensions, each corresponding to one of the four quality aspects) and sort these pairs in ascending order. The top 10 pairs are selected and reported. Between two exam scripts, the less difference indicates they tend to be equivalent in assessment.

\begin{figure*}[ht]
	\centering
	\includegraphics[scale=0.64]{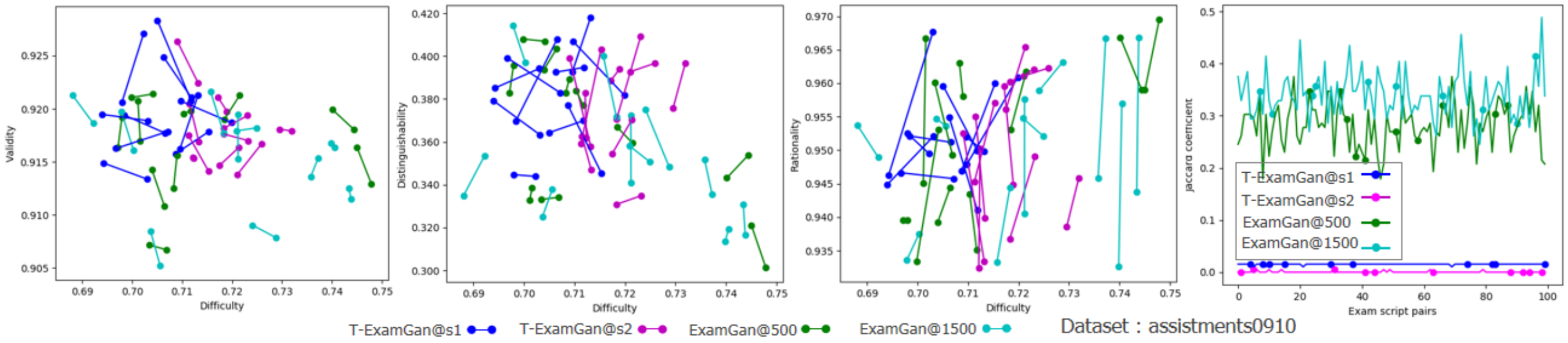}
	\caption{Comparison of exam script pairs on assistments0910}
	\label{fig:ESpair-0910}
 \end{figure*}

\begin{figure*}[ht]
	\centering
	\includegraphics[scale=0.64]{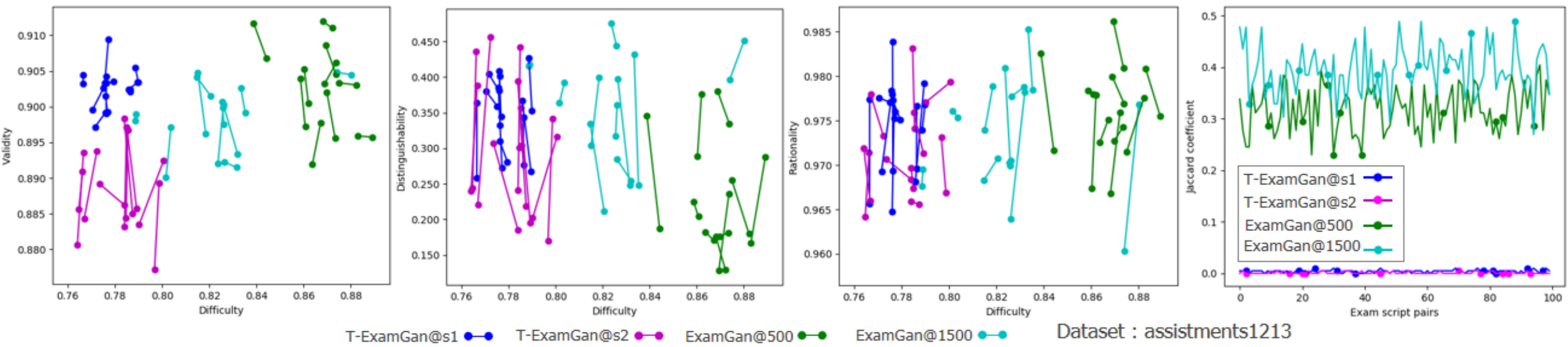}
	\caption{Comparison of exam script pairs on assistments1213}
	\label{fig:ESpair-1213}
 \end{figure*}

\begin{figure*}[ht]
	\centering
	\includegraphics[scale=0.64]{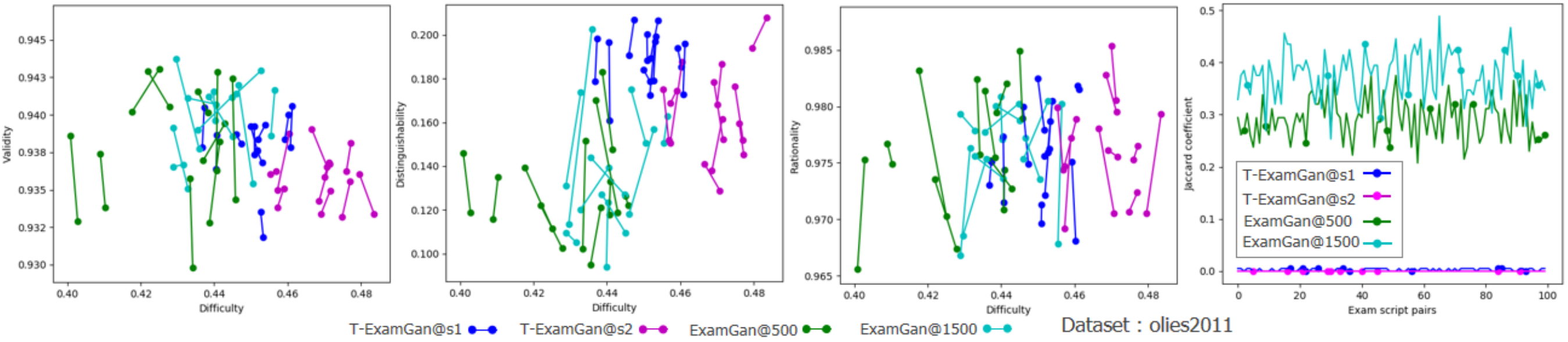}
	\caption{Comparison of exam script pairs on olies2011}
	\label{fig:ESpair-2011}
 \end{figure*}

\textit{Different Question Sets:} The two exam scripts of a pair should have significantly different sets of questions. In Fig. \ref{fig:ESpair-0910}, \ref{fig:ESpair-1213}, and \ref{fig:ESpair-2011} (the right diagram of each figure), the Jaccard coefficient measures to which extent the two exam scripts have different sets of questions by following Equ. (\ref{equ:sim}). Clearly, T-ExamGAN@S1 and T-ExamGAN@S2 ensure the significant difference between the two exam scripts of each pair while ExamGAN@500 and ExamGAN@1500 generate two exam scripts almost same to each other. Also, T-ExamGAN@S2 demonstrates much better performance compared to T-ExamGAN@S1. The T-ExamGAN@S2's attributes is caused by the \textit{twin difference priority} training strategy.     


\textit{Assessment Equivalency:} The two exam scripts of a pair should be equivalent in assessment (i.e., the scores of students are comparable even though they take different exam scripts). Specifically, the two exam scripts should have similar \textit{difficulty}, \textit{distinguishability}, \textit{validity} and \textit{rationality}. In Fig. \ref{fig:ESpair-0910}, \ref{fig:ESpair-1213}, and \ref{fig:ESpair-2011} (the left three diagrams of each figure), the difference between the exam scripts of each pair is illustrated by the length of the link between them. Compared with ExamGAN@500 and ExamGAN@1500, it is believed conceptually that the pairs generated by T-ExamGAN@S1 and T-ExamGAN@S2 tend to have the longer link between them (i.e., greater difference), and such loss in Assessment Equivalency is the cost for the optimization goal, i.e., the two exam scripts in a pair should have different sets of questions. However, it is interesting that the loss in Assessment Equivalency is trivial overall as shown in Fig. \ref{fig:ESpair-0910}, \ref{fig:ESpair-1213}, and \ref{fig:ESpair-2011}. Moreover, we observe that T-ExamGAN@S1 is slightly better than T-ExamGAN@S2. The training strategy applied in T-ExamGAN@S1 enforce high quality of the exam scripts such that the exam scripts tend to share more similar quality.  

\textit{Exam Script Quality:} As shown in Fig. \ref{fig:ESpair-0910}, \ref{fig:ESpair-1213}, and \ref{fig:ESpair-2011} (the left three diagrams in each figure), the quality of individual exam scripts generated by ExamGAN@500 and ExamGAN@1500 does not have noticeable advantage compared to that by T-ExamGAN@S1 and T-ExamGAN@S2. Look closely, the key reason is that the generators and discriminator in T-ExamGAN@S1 and T-ExamGAN@S2 are initialized as those in T-ExamGAN@S2. 

In summary, T-ExamGAN@S1 and T-ExamGAN@S2 demonstrate their unique advantages in generating a pair of high quality exam scripts which are equivalent in assessment and have significantly different sets of questions.  

\section{Case Study}\label{sec:casestudy}
A case study has been done in a real teaching scenario in the School of Computer Science, South China Normal University. The undergraduate course "Programming" is selected. The course has 68 knowledge points and a set of 240 exercises, each exercise contains 1 to 3 knowledge points. The students can do the exercises online. In the case study, we used 533 students’ 384,945 exercise answer records, from 09/2020 to 01/2021, to train a DKT model (AUC of 0.83606). In order to train ExamGAN, following method introduced in Section \ref{sec:brutal}, we randomly generated 100 groups from these 533 students (each with 30 students) to generate the training data. 

We use three real classes of the same course in the new semester from 02/2021 to 04/2021 and create real exam scripts for them at the end of the semester using the trained ExamGAN. For each class, the knowledge mastery level of students is identified based on exercises answered in the semester following Section \ref{sec:firstlayer} and fed to the generator. The exam scripts consists of questions from the exam question bank which consists of the set of exercises and 100 more questions. Each exam script contains 40 questions and the full score is 100, 2.5 for each question. From the test results, \emph{difficulty}, \emph{distinguishability}, \emph{rationality}, and \emph{validity} of the generated exam scripts can be derived and shown in TABLE \ref{tab:casestudy}. Clearly, they are highly desirable and consistent for all classes.

 \begin{table}
 \centering
	\caption{Test result by using the generated exam scripts.}
	\label{tab:casestudy}
	\begin{tabular}{rccc}
	\hline
	 & class-1 & class-2 & class-3 \\ \hline
	 Difficulty & 0.754 & 0.712 & 0.723 \\ \hline
	 Distinguishability & 0.359 & 0.397 & 0.384 \\ \hline
	 Rationality & 0.958 & 0.962 & 0.972 \\ \hline
 	 Validity & 0.934 & 0.929 & 0.940 \\ \hline
	\end{tabular}
  \end{table}

\section{Conclusion and Future Work}\label{sec:conclusion}
To fill the gap in the research field of automatic exam script generation, this paper has delivered the ExamGAN model to generate high quality exam scripts featured in \textit{validity} (i.e., the proper knowledge coverage), \textit{difficulty} (i.e., the expected average score of students), \textit{distinguishability} (i.e., the ability of distinguishing academic performances between students), and \textit{rationality} (i.e., the desirable student score distribution). Interestingly, this study also reveals that ExamGAN can be used to verify whether an update of the exam question bank is necessary to maintain the quality assessment. In addition, ExamGAN has been extended to T-ExamGAN to generate a pair of high quality exam scripts which are equivalent in assessment and have significantly different sets of questions. Based on extensive experiments on three benchmark datasets and a case study in a real teaching scenario, the superiority of the proposed solutions has been verified in various aspects. 

% or


%\appendices

%\section{}

\ifCLASSOPTIONcompsoc
  \section*{Acknowledgments}
\else
  \section*{Acknowledgment}
\fi
This work was supported by the program of National Natural Science Foundation of China No. U1811263; Australia Research Council Linkage Project No.LP180100750, Discovery Project No.DP210100743; and the Foundation of China Scholarship Council No.201808440652.

\ifCLASSOPTIONcaptionsoff
  \newpage
\fi



%\bibliographystyle{IEEEtran}


\bibliographystyle{IEEEtran}
%\bibliography{reference.bib}


%\begin{IEEEbiography}{Zhengyang Wu}
%\end{IEEEbiography}

%\begin{IEEEbiography}{Ke Deng}
%\end{IEEEbiography}

%\begin{IEEEbiography}{Judy Qiu}
%\end{IEEEbiography}

%\newpage
%\begin{IEEEbiography}{Yong Tang}
%\end{IEEEbiography}

%\vfill


\end{document}